\documentclass[10pt,twocolumn,letterpaper]{article}

\usepackage[pagenumbers]{cvpr} 

\usepackage{graphicx}
\usepackage{amsmath}
\usepackage{amssymb}
\usepackage[ruled,vlined]{algorithm2e}
\usepackage{booktabs}

\newcommand{\spara}[1]{\vspace*{0.04in}\noindent\textbf{#1.}\hspace{1mm}}

\usepackage[pagebackref,breaklinks,colorlinks]{hyperref}

\usepackage[capitalize]{cleveref}
\crefname{section}{Sec.}{Secs.}
\Crefname{section}{Section}{Sections}
\Crefname{table}{Table}{Tables}
\crefname{table}{Tab.}{Tabs.}

\def\cvprPaperID{11826} 
\def\confName{CVPR}
\def\confYear{2022}

\begin{document}

\title{Toward Compact Parameter Representations for\\Architecture-Agnostic Neural Network Compression}

\author{Yuezhou Sun$^1$~~~~Wenlong Zhao$^2$~~~~Lijun Zhang$^2$~~~~Xiao Liu$^2$~~~~Hui Guan$^2$~~~~Matei Zaharia$^1$ \\
  $^1$Stanford University~~~~$^2$University of Massachusetts Amherst \\
  \tt\small{\{asunyz, matei\}@cs.stanford.edu}\\
  \tt\small{\{wenlongzhao, lijunzhang, xiaoliu1990, huiguan\}@cs.umass.edu}}
\maketitle

\begin{abstract}
This paper investigates deep neural network (DNN) compression from the perspective of compactly representing and storing trained parameters.
We explore the previously overlooked opportunity of cross-layer architecture-agnostic representation sharing for DNN parameters.
To do this, we decouple feedforward parameters from DNN architectures and leverage additive quantization, an extreme lossy compression method invented for image descriptors, to compactly represent the parameters. 
The representations are then finetuned on task objectives to improve task accuracy. 
We conduct extensive experiments on MobileNet-v2, VGG-11, ResNet-50, Feature Pyramid Networks, and pruned DNNs trained for classification, detection, and segmentation tasks. 
The conceptually simple scheme consistently outperforms iterative unstructured pruning. Applied to ResNet-50 with 76.1\% top-1 accuracy on the ILSVRC12 classification challenge, it achieves a $7.2\times$ compression ratio with no accuracy loss and a $15.3\times$ compression ratio at 74.79\% accuracy.
Further analyses suggest that representation sharing can frequently happen across network layers and that learning shared representations for an entire DNN can achieve better accuracy at the same compression ratio than compressing the model as multiple separate parts.
We release PyTorch code to facilitate DNN deployment on resource-constrained devices and spur future research on efficient representations and storage of DNN parameters.
\end{abstract}

\section{Introduction}
\label{introduction}
Deep neural networks (DNNs) have achieved state-of-the-art performances on many computer vision tasks. 
Compared to serving models on clouds, deploying DNNs on mobile and IoT devices benefits from better privacy, lower latency, and stronger security~\cite{shi2016edge}. On-device inference has enabled many emerging applications such as autonomous driving, social robots, virtual reality, and wearable devices. 

A major roadblock to the on-device deployment of DNNs is their large model sizes. 
State-of-the-art DNNs usually contain millions of parameters. 
A ResNet-50 based Feature Pyramid Network (FPN) \cite{lin2017feature}, for example, contains 41.7 million parameters and requires 167MB to store. 
The requirements quickly add up to the scale of GB when we intend to host tens of applications on a device, each involving one or more DNN-based inference tasks. 
In reality, however, edge devices are equipped with limited and often insufficient processor memory and storage space for DNNs due to technical and economical factors. 
Microcontroller units (MCUs) and low-end IoT devices typically have hundreds of KB to hundreds of MB of SRAM~\cite{lee2020fast, lin2021mcunetv2}. 
Even on mobile devices that have relatively large storage space, chip memories are still constrained. Further, the size of apps that can be transmitted to edge devices is often restricted by system regulations\footnote{To cite an example, iOS12 prevents users from downloading apps of more than 200MB through cellular data.} and constrained network bandwidth in many regions in the world.

Recent years have witnessed dramatic progress in developing DNN compression techniques to reduce DNN model sizes. 
Pruning \cite{gale2019state, blalock2020state, frankle2018lottery, ma2021sanity}, distillation \cite{hinton2015distilling}, and light-weight architecture search and design methods \cite{iandola2016squeezenet, tan2019mnasnet} explore diverse venues toward identifying sparse or small architectures and training their associated parameters to achieve good task performances. 
Low precision methods~\cite{wang2019haq, rastegari2016xnor, zhou2016dorefa, zhou2017incremental, choi2018pact, guan2019post, jin2020adabits, zhang2018lq}, on the other hand, reduce the bit-width of each parameter for space-efficiency.
This rich literature commonly preserves the one-to-one mappings between parameters and architectures in the compressed models and does not exploit intra-layer or cross-layer redundancy among parameters to reduce model sizes.
Previous works on applying quantization for DNN compression break the one-to-one mappings to some extent by clustering parameters in a specific layer and allowing each cluster to share representations~\cite{han2015deep, gong2014compressing, wu2016quantized, son2018clustering, stock2019and, martinez2021permute}. None of the existing works, however, investigates the opportunity of architecture-agnostic cross-layer representation sharing to achieve high space-efficiency and general applicability. 

This paper studies the previously overlooked opportunity of cross-layer parameter representation sharing to compress DNNs. It highlights the possibility to compactly represent and store trained DNN parameters, independently of their network structures. 
The key observation is that parameters, once decoupled from model architectures, are unstructured data that can possibly be compressed by any general compression methods. Lossy compression algorithms, for example, have achieved appealing size-quality trade-offs at multimedia data \cite{torralba2008small, babenko2014additive}. But their potential to compress feedforward parameters from DNNs is largely overlooked.
We thus aim to unveil the possibility of significantly compressing DNN parameters independently of their network architectures with no or negligible task performance loss.

To this end, we propose AQCompress, an architecture-agnostic DNN compression scheme, that leverages additive quantization (AQ)~\cite{babenko2014additive}, an extreme lossy compression method invented for image descriptors, to compactly represent trained DNN parameters from any network layers. 
The compressed representations, referred to as \textit{code representations} from now on, consist of discrete codes and codebooks which capture shareable local patterns across network layers. 
We manipulate the code distribution learned by AQCompress to be more skewed so that Huffman coding of the codes can achieve high space-efficiency.
AQCompress further features a gradient propagation mechanism which finetunes the codebook parameters on task objectives. 
During inference, the compact representations can easily reconstruct DNN parameters through fast addition operations. We select AQ among many data compression methods in our investigation due to this efficient decoding.

To validate the effectiveness and generality of AQCompress, we evaluate it on MobileNet-v2 \cite{sandler2018mobilenetv2}, VGG-11 \cite{simonyan2014very}, ResNet-50 \cite{he2016deep}, and pruned DNNs trained for CIFAR-10, CIFAR-100 \cite{krizhevsky2009learning}, and ImageNet \cite{imagenet_cvpr09} classification tasks. 
We also apply it to FPN models \cite{lin2017feature} trained for COCO object detection, keypoint detection, and panoptic segmentation tasks \cite{lin2014microsoft}. 
AQCompress consistently outperforms or achieves competitive performances compared to recently proposed pruning and quantization methods \cite{frankle2018lottery, liu2018rethinking, renda2020comparing, han2015deep, wang2019haq}. Applied to ResNet-50 with 76.1\% top-1 accuracy on the ILSVRC12 classification challenge, it achieves a $7.2\times$ compression ratio with no accuracy loss and a $15.3\times$ compression ratio at 74.79\% accuracy. 

Our ablation study shows that, at the same compression ratio, learning shared representations for all parameters from a MobileNet-v2 trained for CIFAR-100 achieves better accuracy than compressing it as 4, 8, or 16 separate parts. This confirms the intuition that allowing more parameters to share representations can lead to better size-accuracy trade-offs. 
Further analyses verify that sharing can frequently happen across network layers in the learned parameter representations. These are an unconventional observation. 
While the recent DNN compression literature largely emphasizes the design, search, and training of lightweight network architectures, our results indicate that future lines of research focusing on compact representations and storage of trained DNN parameters and their combination with light-weight DNN architectures may lead to new possibilities for on-device AI deployment.


\begin{figure*}[t]
  \centering
  \includegraphics[width=0.99\textwidth]{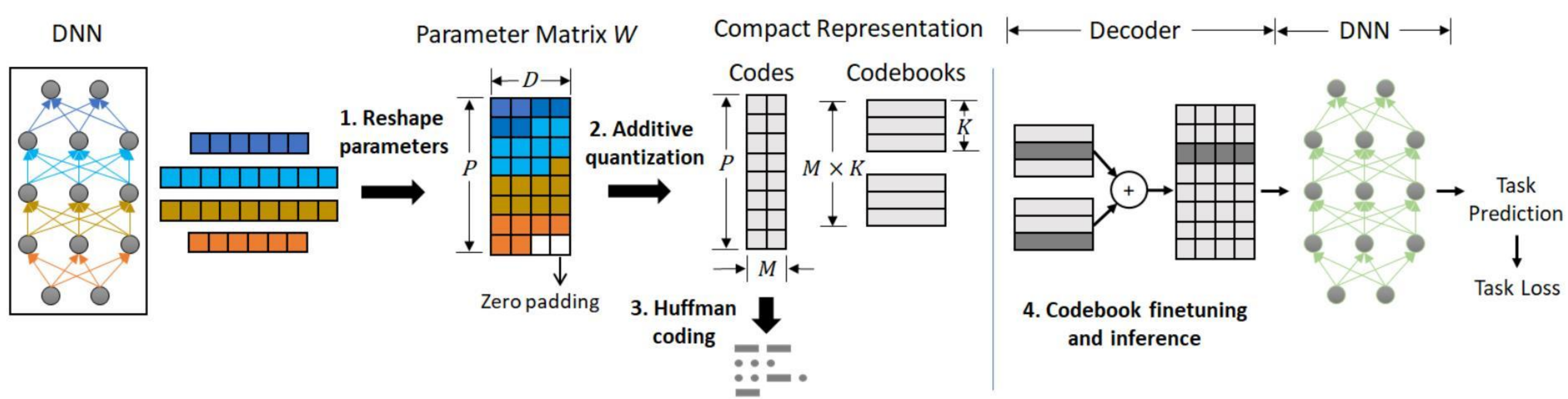}
  \caption{The compression pipeline of AQCompress.}
  \label{fig:pipeline}
\end{figure*}

\section{Related Work} \label{related}

The literature on developing space-efficient DNN representations can broadly be categorized into pruning, quantization, light-weight DNN architecture design, knowledge distillation.
Below, we introduce the work most relevant to ours. Readers can refer to recent surveys~\cite{gale2019state, blalock2020state, qin2020binary, gholami2021survey, chen2019deep, liu2020bringing} for comprehensive reviews.

\spara{Pruning}
Network pruning aims to remove connections based on some importance criteria to reduce model sizes or computational complexity~\cite{lecun1990optimal}. Existing approaches are either structured or unstructured pruning. Structured pruning removes parameters in groups, such as filters or channels, and thus maintains the regular patterns in network structures~\cite{li2016pruning}. Unstructured pruning, on the contrary, removes individual weights in DNNs and can achieve significantly greater compression than structured pruning~\cite{han2015learning, frankle2018lottery}. In this paper, we compare AQCompress with iterative unstructured pruning \cite{renda2020comparing} to demonstrate the high compression ratios attained by AQCompress. We also show that AQCompress remains effective on parameters from pruned networks and thus complements pruning methods.

\spara{Quantization}
Quantization methods can be classified into scalar and vector quantization. Scalar quantization (SQ)~\cite{wang2019haq, rastegari2016xnor, zhou2016dorefa, zhou2017incremental, choi2018pact, agustsson2017soft} represents each parameter with less bits. To compensate for accuracy loss due to the low bit-width, SQ usually requires low-precision training~\cite{courbariaux2014training, li2016ternary, mcdonnell2018training, lin2017towards, abdolrashidi2021pareto, zhang2018lq, jacob2018quantization}, which is prone to convergence issues. While 8-bit quantization (i.e. 4$\times$ compression) is usually robust, vector quantization (VQ) \cite{han2015deep, gong2014compressing, stock2019and, carreira2017model} is developed to achieve higher compression ratios.

VQ clusters vectors into groups and represents each group by a centroid. Each vector is then represented by the index of its closest centroid. 
Product quantization (PQ) \cite{jegou2010product} decomposes each vector into components in orthogonal subspaces, and then applies VQ to the components.
Additive quantization (AQ)~\cite{babenko2014additive} generalizes PQ and represents each vector as a sum of several components, but does not involve space decomposition and thus is exempted from the orthogonal subspace independence assumptions. AQ is also easy to decode since each original vector can be reconstructed by summing component vectors.

Few works have compressed DNN layers using vector quantization techniques. 
Gong \etal~\cite{gong2014compressing} applies VQ to compress fully connected layers. 
Wu \etal~\cite{wu2016quantized} shows that minimizing activation errors gives better accuracy than minimizing parameter quantization errors. 
Son \etal \cite{son2018clustering} leads the exploration of quantizing 3$\times$3 convolutional layers and finetuning the centroids to encode filter rotation. 
Stock \etal~\cite{stock2019and} minimizes output errors of DNN layers sequentially and applies distillation for finetuning. 
Chen \etal \cite{chen2020towards} further minimizes parameter reconstruction errors and task errors jointly. Martinez \etal~\cite{martinez2021permute} permutes parameters to make a DNN layer easier to quantize. Orthogonal to the above methods are mixed-precision methods, which groups scalars or vectors by sensitivity to quantization and separately use high and low bit widths for them.

Different from existing studies, our work investigates the possibility of compressing model parameters independently of network architectures. 
We aim to highlight the value of previously unexplored opportunities for learning compact parameter representations shared across layers of possibly diverse shapes. 
Although beyond the goal of this paper, exciting follow-up work can explore the combination of alternative training objectives \cite{chen2020towards, stock2019and}, permutation approaches \cite{martinez2021permute}, and mix-precision methods with AQCompress.
To our best knowledge, this work is also the first to validate the potential of AQ on compressing feedforward parameters from DNNs. 

\spara{Light-Weight Architecture Design} 
Light-weight DNN design, including developing compact student models in knowledge distillation and multitask models,
is specific to tasks and results in limited compression ratios. Light-weight architectures such as MobileNet~\cite{sandler2018mobilenetv2} and MnasNet~\cite{tan2019mnasnet} are either manually crafted with expertise or identified by neural architecture search techniques that involve long tuning processes. 
On the other hand, there is no established principle for how to design and train student architectures in distillation methods and multitask models.
Our experiments show that AQCompress can effectively compress a wide variety of models including MobileNet-v2, which is commonly considered a light-weight DNN. AQCompress can technically also be applied to compress any other light-weight DNN models.




\section{Compact Parameter Representations for Architecture-Agnostic DNN Compression} 
Given trained DNNs, our goal is to compactly represent their parameters independently of the network architectures. In this section, we introduce and motivate the design choices for building our proposed architecture-agnostic DNN compression scheme, AQCompress, to learn compact representations for trained DNN parameters.

\subsection{Overview}
Our compression pipeline has four steps (\cref{fig:pipeline}). 
First, it decouples trained parameters from DNN architectures and reshapes them into a parameter matrix, to which general data compression approaches can be applied. 
Second, it learns compact code representations for the parameters through an encoder-decoder structure. The code representations consist of discrete codes and codebooks.
Third, it manipulates the code distribution to be more skewed and leverages Huffman coding to compactly represent the codes. 
Finally, it finetunes the codebooks, i.e. trainable parameters in the decoder, on task objective functions to improve task performances. 

\subsection{Composing Parameter Matrix}
Additive quantization, like other lossy compression methods for multi-media data, is invented for data without network structures. 
The first step of our compression pipeline is to rather boldly decouple trained DNN parameters from their associated architectures and consider the parameters as a piece of unstructured data.
Given a trained DNN, AQCompress extracts its parameters and reshapes them into a parameter matrix $W\in\mathbb{R}^{P\times D}$, where $P,D\in\mathbb{Z}^+$ are hyperparameters and each $D$-dimensional row vectors can be called a \textit{parameter page}\footnote{Named after ``memory page", the smallest unit of data for memory management in operation systems.}. 
We compose matrix $W$ with parameters from all layers of a DNN in order that the compression algorithm can learn shareable patterns among parameters across DNN layers. Assume the DNN has $T$ scalar parameters.  The number of parameter pages is $P = \lceil \frac{T}{D} \rceil$, where $\lceil~\rceil$ denotes rounding up to the nearest integer. Zero-padding is used in case $\frac{T}{D}\notin\mathbb{Z}$.


\subsection{Learning Compositional Representations}\label{recon}
AQCompress compactly represents the parameter matrix $W$ by the additive quantization coding scheme (AQ) \cite{babenko2014additive, codebooks}. 
We select AQ among many data compression methods in our investigation due to its efficient decoding: the compact representations can easily reconstruct DNN parameters through fast addition operations during inference.

The compact representations are learned through an encoder-decoder structure.
The learned representations consist of $P$ sets of \textit{discrete codes}, one for each parameter page, and $M$ shared \textit{codebooks}, each with $K$ \textit{basis vectors}, where $M$ and $K$ are hyperparameters. The codebooks and codes can be used to reconstruct DNN parameters through efficient addition operations.

\spara{Encoder} Let $w_p\in\mathbb{R}^D$ be the $p$-th parameter page, i.e. the $p$-th row vector in $W$. The encoder inputs $w_p$ and projects it into a set of $M$ activations
\begin{equation}
\alpha_p^m = \text{softplus}(\theta_2^{m\top}\tanh(\theta_1^{m\top}w_p+b_1^m)+b_2^m) \in\mathbb{R}^K
\end{equation}
for $m\in [0, M)\cap\mathbb{Z}$, where $\theta_1^m\in\mathbb{R}^{D\times H}$, $b_1^m\in\mathbb{R}^H$, $\theta_2^m\in\mathbb{R}^{H\times K}$, and $b_2^m\in\mathbb{R}^K$ are the trainable parameters. 
Each activation vector $\alpha_p^m$ is then converted into a probability distribution $d_p^m\in\mathbb{R}^K$ that approximates a one-hot vector by the Gumbel-softmax parameterization trick \cite{jang2016categorical, maddison2016concrete} with temperature $\tau=1$. 

\spara{Decoder} Let $A^m\in\mathbb{R}^{K\times D}$ be the $m$-th codebook containing $K$ basis vectors of $D$ dimensions, where $m\in[0,M)\cap\mathbb{Z}$. The basis vectors are aggregated according to the distributions $d_p^m$ to reconstruct $w_p$ as
\begin{equation}\label{eq:recon}
\tilde{w}_p = \sum^{M-1}_{m=0}A^{m\top}d_p^m\in\mathbb{R}^D.
\end{equation}
We train codebook parameters $A$ with
the encoder parameters to minimize an $L_2$ parameter page reconstruction loss
\begin{equation}
\mathcal{L}_{\text{reconstruction}}(A, \theta_1, b_1, \theta_2, b_2)=\frac{1}{P}\sum_{p=0}^{P-1}||w_p-\tilde{w}_p||^2.
\end{equation}

After training the encoder-decoder, each discrete code $\arg\max d_p^m$ can serve as an index for the basis vector to select from the $m$-th codebook. $M$ selected basis vectors, one per codebook, are summed to reconstruct the original parameter page $w_p$ as in \cref{eq:recon}.

\subsection{Code Representations and Space-Efficiency}\label{coderep}
Given a parameter page $w_p$, we can pass it through the encoder structure and record $\arg\max d_p^m$ for $m\in[0,M)\cap\mathbb{Z}$. This results in $M$ discrete codes that can compactly represent $w_p$. Along with the codebooks $A$ shared by all parameter pages, these codes can effectively reconstruct the original $w_p$ as in \cref{eq:recon}. We use Huffman coding to compactly store the discrete codes.

Huffman coding is a classic algorithm developed to find an optimal prefix tree for lossless data compression \cite{huffman1952method}. Directly applying Huffman coding to compress codes results in limited compression because the values in the discrete codes are randomly distributed. To achieve high compression ratios with Huffman coding, we sort the basis vectors in each codebook by frequency so that the basis vectors used more often correspond to smaller codes. Consequently, in the code representations of a DNN, smaller discrete codes appear strictly more often than larger ones. The code distribution is thus manipulated to be more skewed.

We compare Huffman coding with bit-wise representations, which store each code using the least number of bits possible and is detailed in the Appendix. We show in Experiment \cref{subsection:ablation} that Huffman coding without sorting basis vectors is more space-efficient than the bit-wise representations, while manipulating code distributions significantly improves the compression ratios as compared to no sorting.



\subsection{Codebook Finetuning}
We finetune the codebook parameters $A$, i.e. the decoder for learning the code representations, on task objective functions in order to achieve better task performances. We fix the discrete codes during finetuning for simplicity. 

The intuitive approach to implement finetuning is to construct a decoder computational graph, where codebook parameters are trained to reconstruct DNN parameters, and a task DNN computational graph, where the produced DNN parameters are mapped to the task DNN architecture as trainable parameters. Common deep learning frameworks, such as PyTorch and TensorFlow, do not support gradient propagation across computational graphs. We thus customize the learning mechanism so that gradients from task losses can be back-propagated through the task DNN into the codebook computational graph.

\spara{Forward propagation} 
To construct a parameter page $\tilde{w}_p$ in a DNN, we expand its corresponding $M$-dimensional code representation into $M$ one-hot vectors $d_p^m$. We reconstruct the parameter page according to \cref{eq:recon} and assign it to the task DNN computational graph. Standard feedforward calculations can then be performed.

\spara{Back-propagation}
Let $\mathcal{L}_{\text{task}}$ be a task loss. We calculate the gradients of $\mathcal{L}_{\text{task}}$ with respect to trainable parameters in the task DNN normally. We then detach the gradients from the DNN computational graph and reshape them into a matrix $G\in\mathbb{R}^{P\times D}$, adding zero padding as needed. Let $g_p$ be the $p$-th row vector in the gradient matrix $G$. We have
\begin{equation}\label{task_gradient}
g_p = \frac{\partial\mathcal{L}_{\text{task}}}{\partial \tilde{w_p}} \in \mathbb{R}^D.
\end{equation}

We train the codebooks $A$ to minimize the finetuning loss
\begin{equation}\label{ft_loss}
\mathcal{L}_{\text{finetuning}}(A) = \sum_{p=0}^{P-1} g_p\tilde{w}_p,
\end{equation}
where $\forall p$, $\tilde{w}_p = \sum^{M-1}_{m=0}A^{m\top}d_p^m \in \mathbb{R}^D$ as in the forward propagation. The gradients of the constructed finetuning loss with respect to the codebook basis vectors are the same as those of the task loss as derived in the Appendix.

\begin{figure*}[t]
    \centering
    \begin{subfigure}{.33\textwidth}
        \centering
        \includegraphics[width=\textwidth]{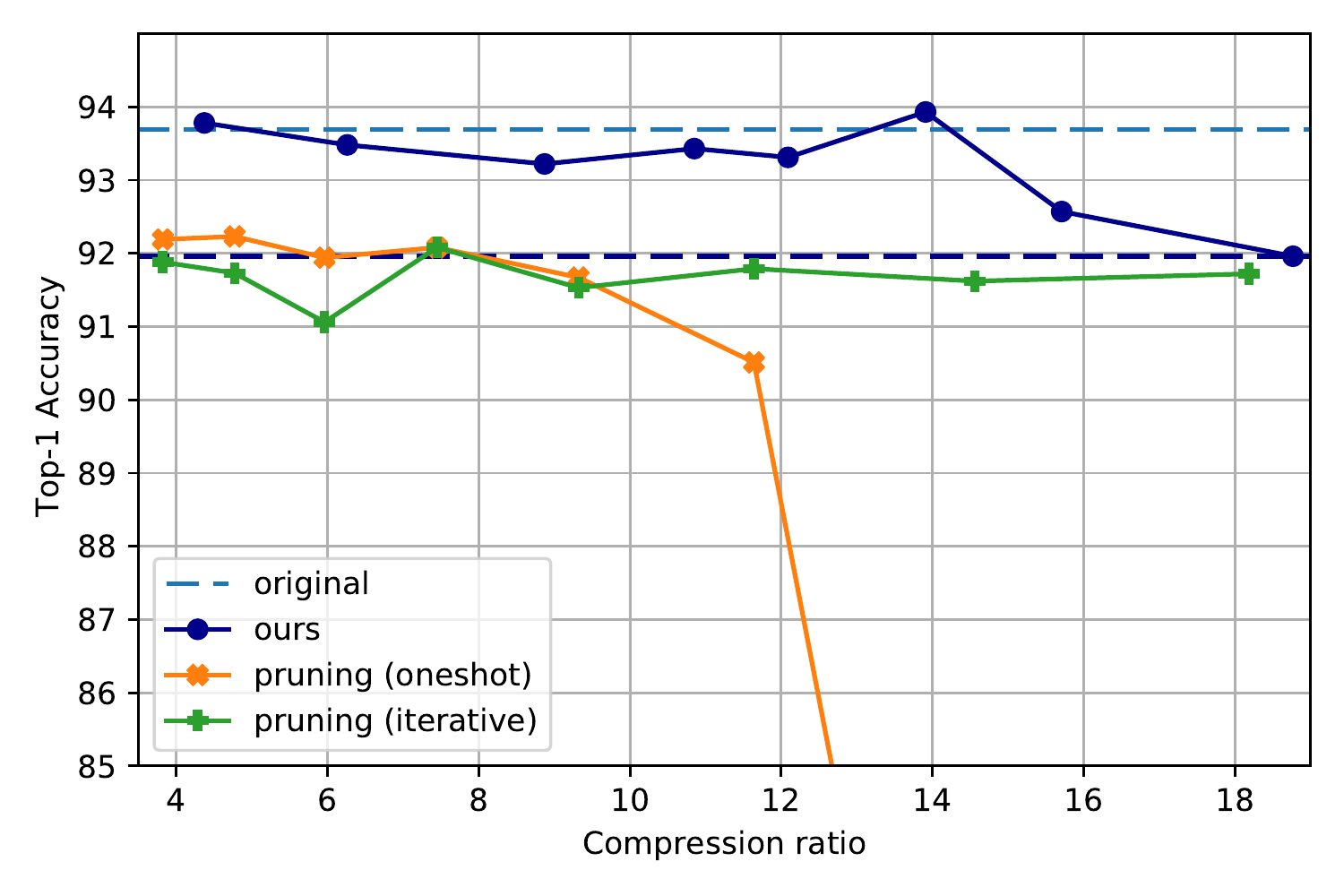}
        \caption{MobileNet-v2 for CIFAR-10.}
    \end{subfigure}\hspace{-0.5em}
    \begin{subfigure}{.33\textwidth}
        \centering
        \includegraphics[width=\textwidth]{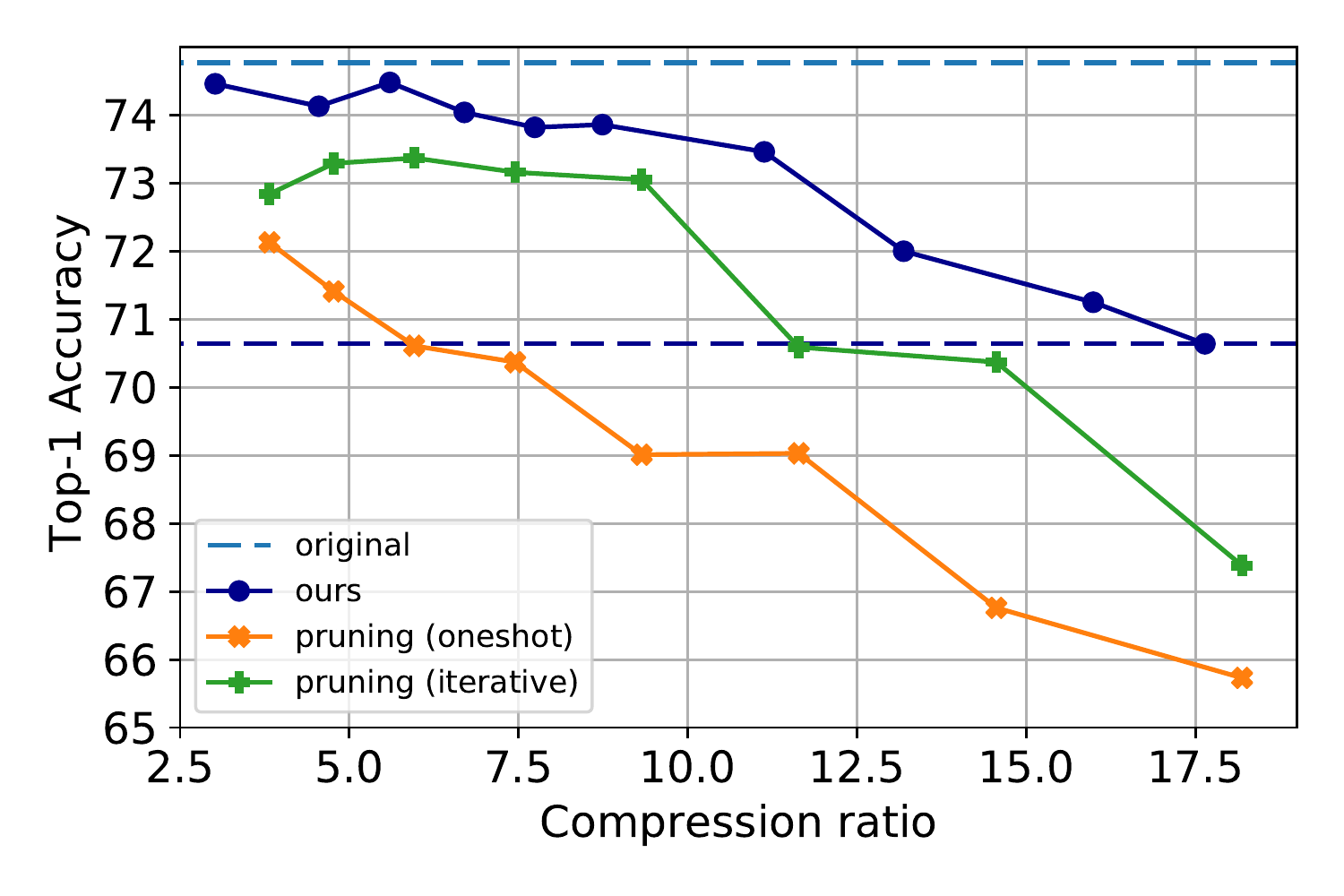}
        \caption{MobileNet-v2 for CIFAR-100.}
    \end{subfigure}\hspace{-0.5em}
    \begin{subfigure}{.33\textwidth}
        \centering
    \includegraphics[width=\textwidth]{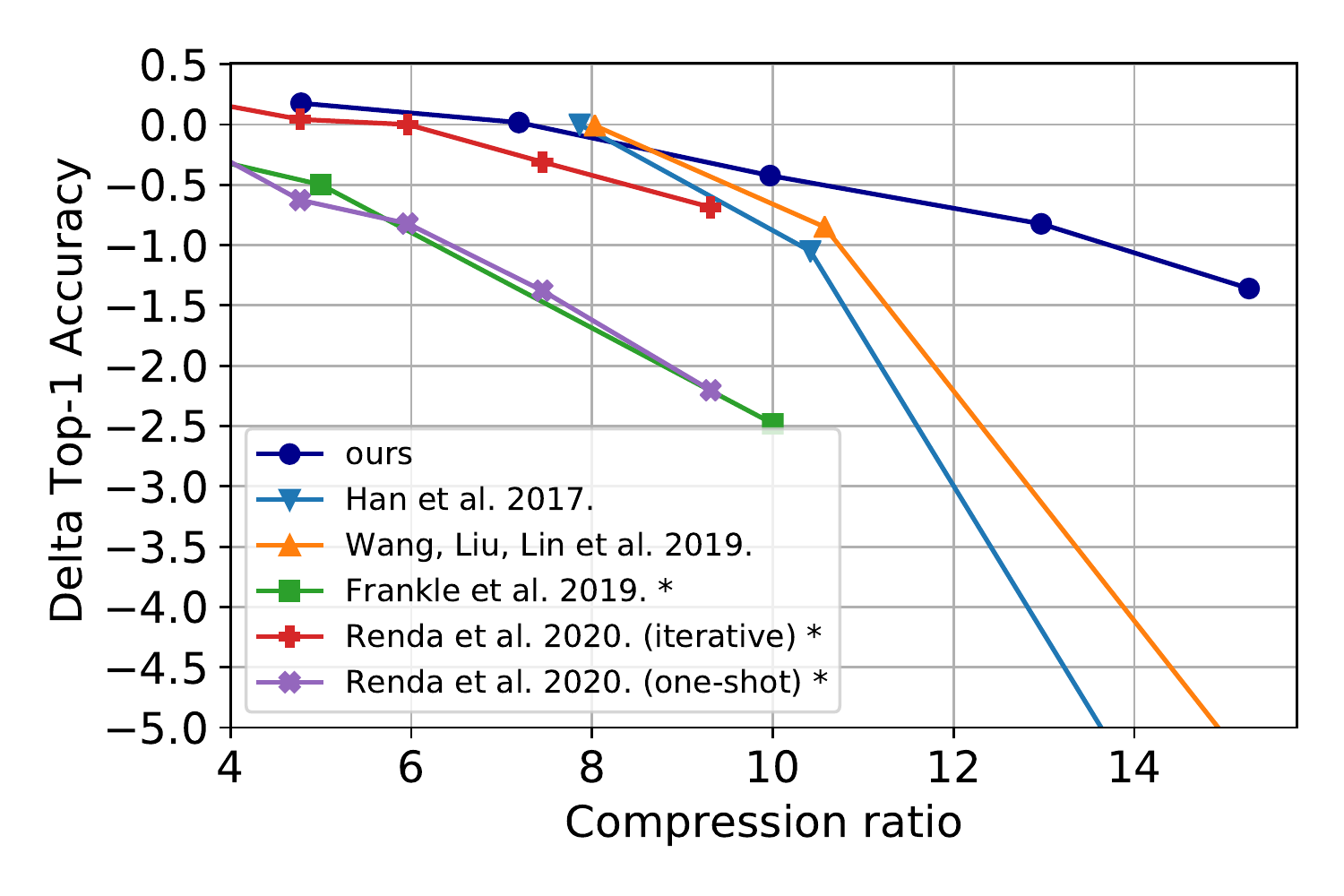}
        \caption{ResNet-50 for ILSVRC-2012. Pruning methods marked by *. Other baselines involve quantization.}
    \end{subfigure}
    \caption{Results for image classification models. In (a) and (b), we implement “one-shot pruning” and “iterative pruning with learning rate rewinding” in Renda et al.\cite{renda2020comparing} as baselines. In (c), we use results reported by the referenced papers as baselines. We report delta accuracy in (c) because the literature involves different pretrained checkpoints with varying accuracies before compression. In all experiments, AQCompress outperforms or achieves competitive performances compared to the baselines.}
    \label{fig:cls}
\end{figure*}

\begin{figure*}[t]
    \centering\hspace{-2em}
    \begin{subfigure}{.33\textwidth}
        \centering
        \includegraphics[width=\textwidth]{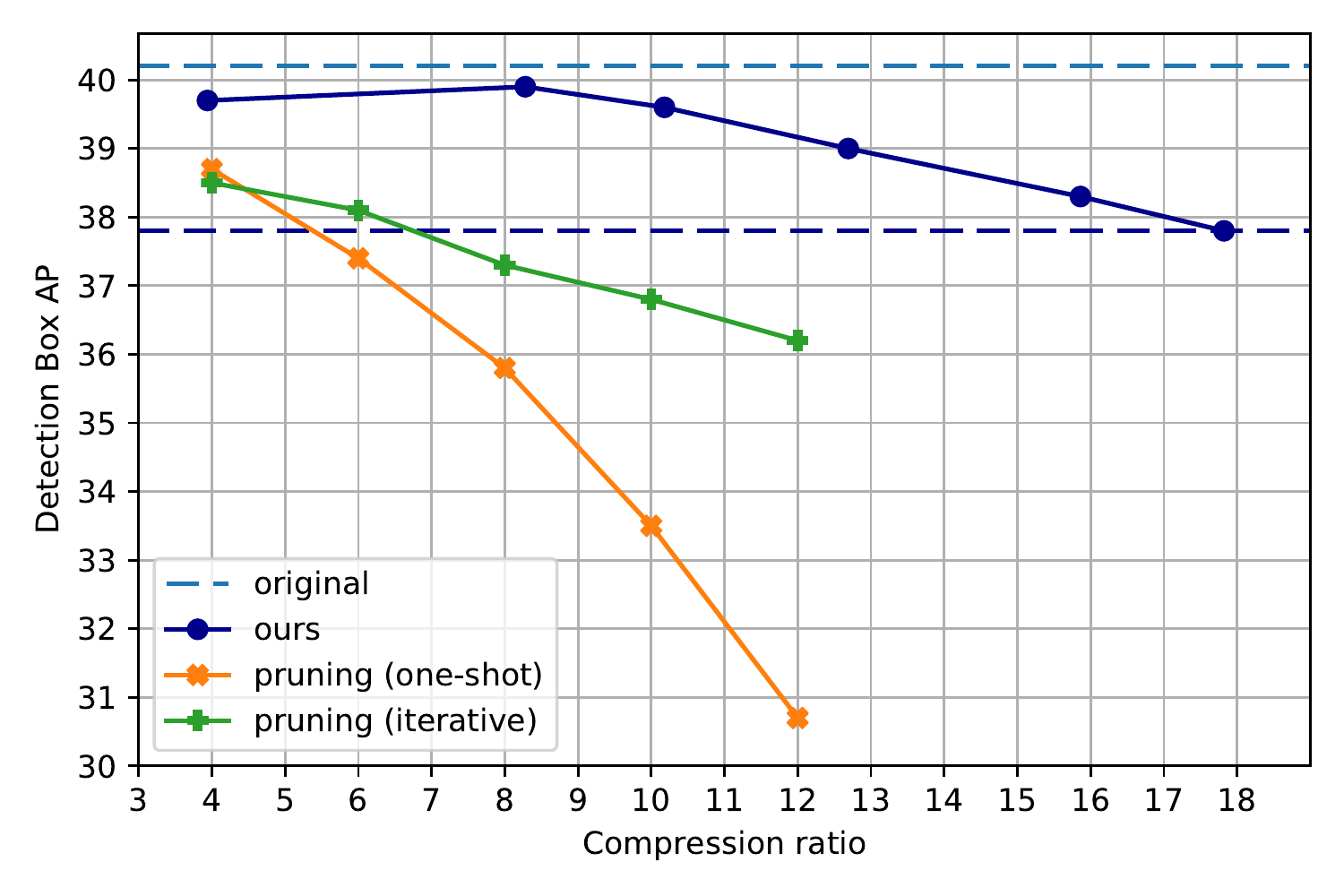}
        \caption{Detection model: box AP}
    \end{subfigure}\hspace{-0.5em}
    \begin{subfigure}{.33\textwidth}
        \centering
        \includegraphics[width=\textwidth]{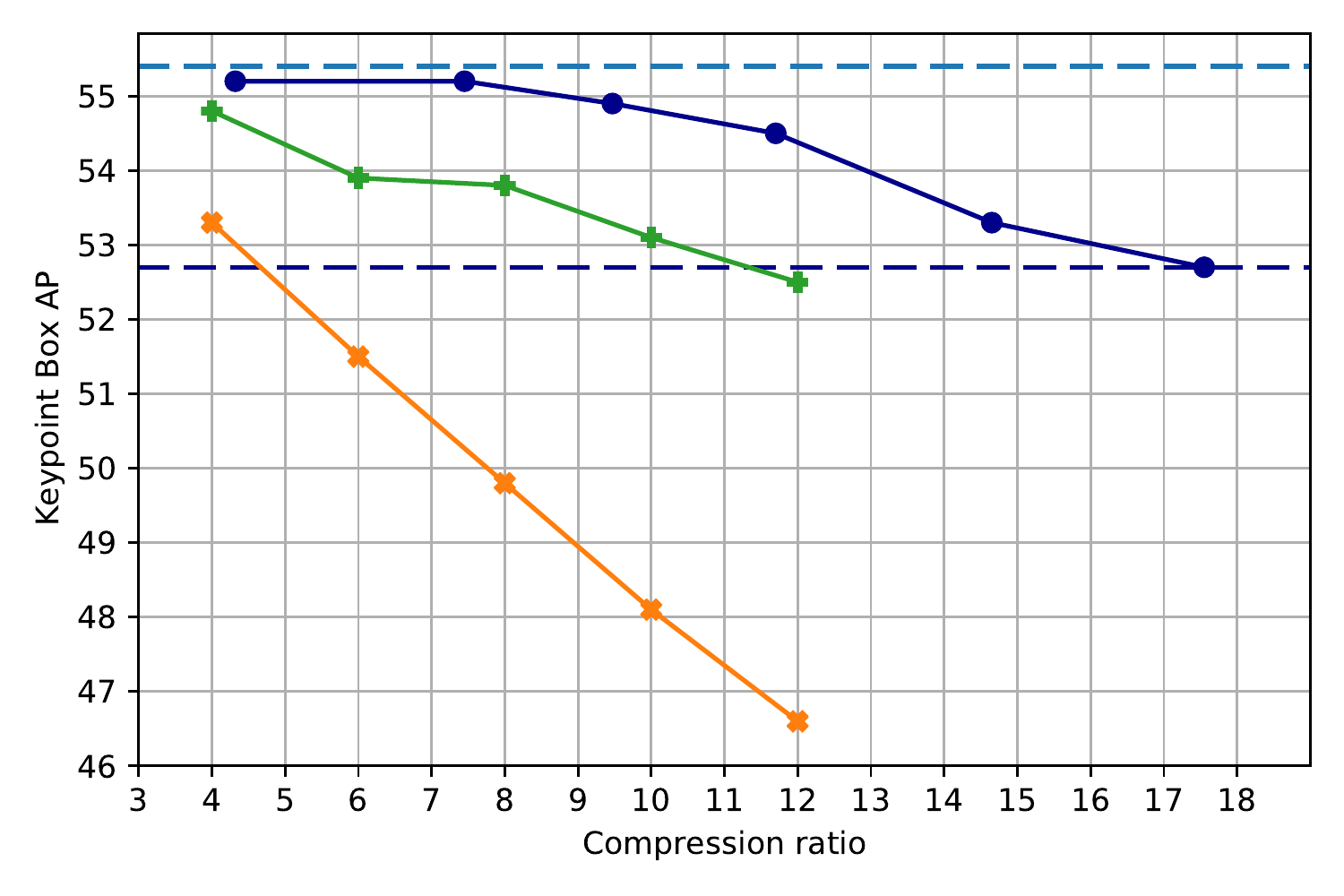} 
        \caption{Keypoint model: box AP}
    \end{subfigure}\hspace{-0.5em}
    \begin{subfigure}{.33\textwidth}
        \centering
        \includegraphics[width=\textwidth]{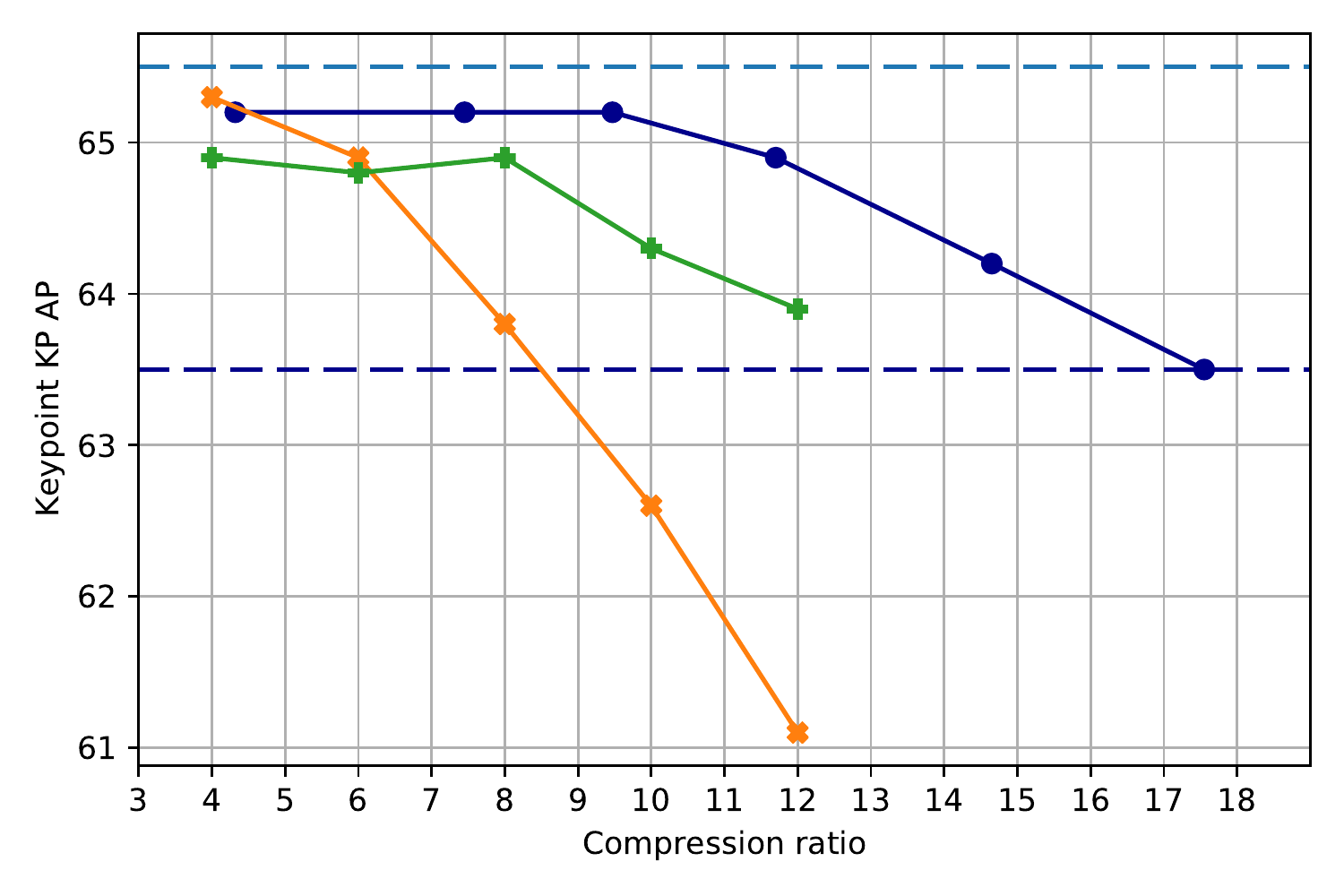}
        \caption{Keypoint model: keypoint AP}
    \end{subfigure} \bigskip \hspace{-2em}
    \begin{subfigure}{.33\textwidth}
        \centering
        \includegraphics[width=\textwidth]{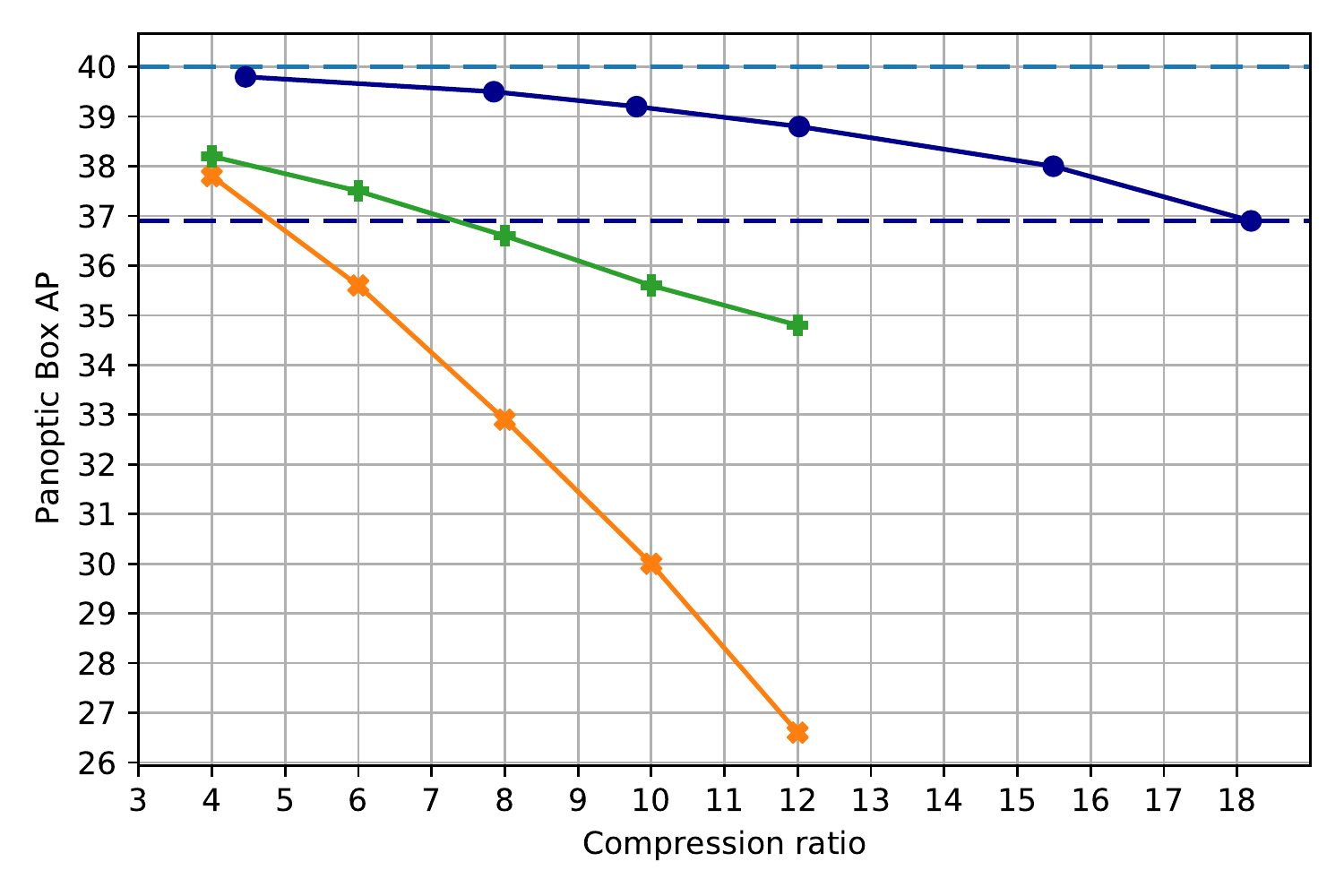}
        \caption{Panoptic model: detection AP}
    \end{subfigure}\hspace{-.5em}
    \begin{subfigure}{.33\textwidth}
        \centering
        \includegraphics[width=\textwidth]{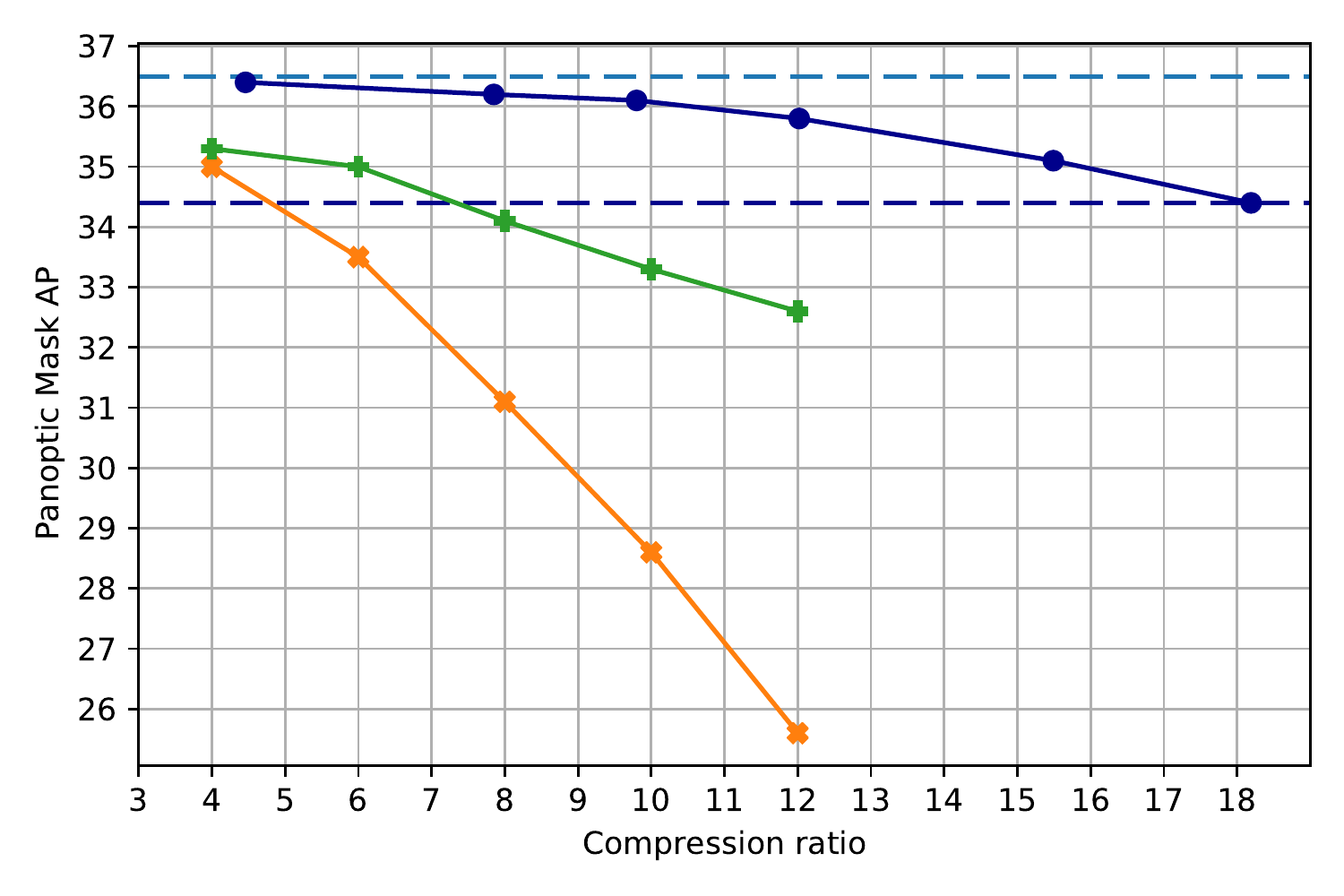}
        \caption{Panoptic model: mask AP}
    \end{subfigure}\hspace{-.5em}
    \begin{subfigure}{.33\textwidth}
        \centering
        \includegraphics[width=\textwidth]{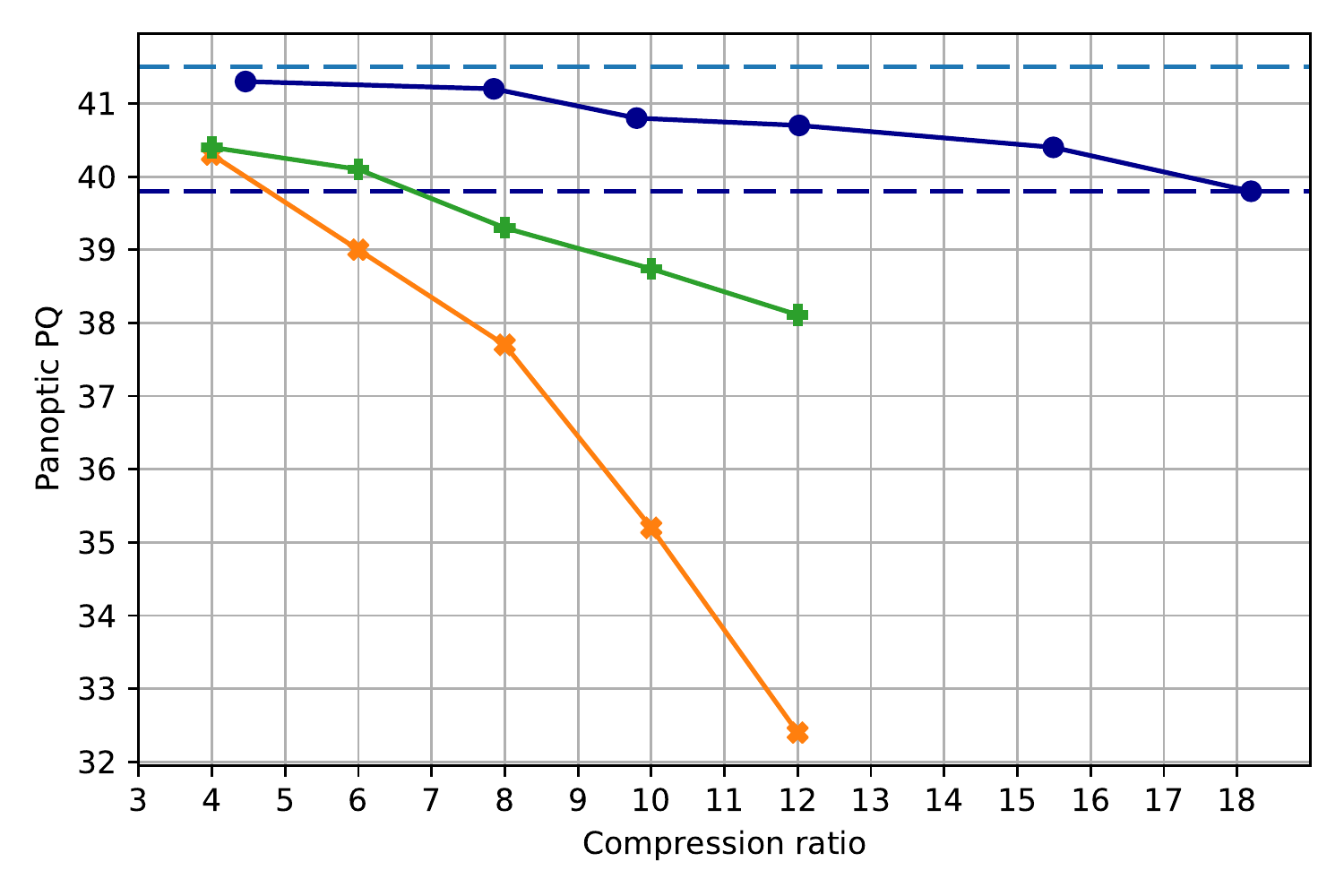}
        \caption{Panoptic model: PQ}
    \end{subfigure}
      \vspace{-0.15in}
    \caption{Compression results for three DNN models from the Detectron2 library on different tasks. Multiple evaluation metrics are plotted. 
    Our method significantly outperforms pruning baselines on all metrics and tasks at high compression ratios.}
    \label{fig:detectron}
\end{figure*}

\section{Experiments}
We conduct extensive experiments to validate the effectiveness and generality of AQCompress and the importance of its components.
All experiments are implemented in Python 3.6 with PyTorch 1.7.1 and run on single Nvidia Titan X, 1080 Ti, or 2080 Ti GPUs. Detailed hyperparameter settings are described in Appendix. 

Many of our experiments compare AQCompress to unstructured pruning \cite{renda2020comparing}, which is a well-studied and widely adopted DNN compression method. We note that the comparisons throughout this paper are in favor of the pruning methods. When computing compression ratios achieved by AQCompress, we account for both codebook and discrete code sizes. However, we approximate the compression ratios of pruning by only considering the remaining parameters but not sparse indices.
During deployment of pruning methods, storing sparse indices in CSR or alternative formats always introduce non-trivial extra overhead.
The compression ratios for pruning are thus over-estimated. Nevertheless, AQCompress consistently outperforms pruning baselines in our experiments.

\subsection{Image Classification} \label{subsection:cls}
\spara{CIFAR-10 and CIFAR-100} We obtain the CIFAR datasets \cite{krizhevsky2009learning} from Torchvision\footnote{https://pytorch.org/vision/stable/datasets.html} and train MobileNet-v2 \cite{sandler2018mobilenetv2} and VGG-11 \cite{simonyan2014very} on them. For evaluation, we report top-1 accuracy, i.e. the percentage of correct classification predictions, on the test set. 
Since Renda \etal~\cite{renda2020comparing} only reports ResNet results (see the paragraph below) on image classification,
we implement one-shot and iterative unstructured magnitude pruning baselines \cite{renda2020comparing, frankle2018lottery} using PyTorch's pruning API\footnote{\url{https://pytorch.org/tutorials/intermediate/pruning_tutorial.html}}. In iterative pruning, we rewind learning rates to the initial status before each finetuning round. 

The MobileNet-v2 results are shown in \cref{fig:cls} (a) and (b); VGG-11 results show similar trends and are in the Appendix. Our method is able to compress MobileNet-v2 on CIFAR-10 to 14$\times$ without accuracy loss, and MobileNet-v2 on CIFAR-100 to 11$\times$ with barely 1.31\% accuracy loss. It also consistently outperforms pruning methods at all compression ratios.

\spara{ImageNet} We obtain the ILSVRC-2012 classification dataset \cite{imagenet_cvpr09} from the official website\footnote{\url{https://www.image-net.org/challenges/LSVRC/2012/index.php}}. We obtain pretrained ResNet-50 \cite{he2016deep} for ImageNet classification from PyTorch Hub\footnote{\url{https://pytorch.org/hub/pytorch_vision_resnet/}}. For evaluation, we report top-1 accuracy on the validation set due to the absence of an official testing set. 
We compare AQCompress with state-of-the-art unstructured pruning \cite{frankle2018lottery, liu2018rethinking, renda2020comparing} and quantization \cite{han2015deep, wang2019haq} approaches that are designed to compress DNNs. 
\cref{fig:cls} shows the trade-offs between the compression ratio and top-1 accuracy. 
AQCompress achieves competitive results compared to existing methods and much less performance loss when the compression ratio is higher than 10$\times$. 

We will show in \cref{subsection:pr+aq} that AQCompress  remains effective on pruned DNNs and also complements pruning.

\subsection{Detectron2: Object Detection, Keypoint Detection, Panoptic Segmentation} \label{subsection:detectron}
The second set of experiments is conducted on the COCO Detection, Keypoint Detection, and Panoptic Segmentation tasks \cite{lin2014microsoft}. 
We download the COCO 2017 dataset from its official website\footnote{\url{https://cocodataset.org}}. The three tasks involve annotated train-validation splits of sizes (80k, 40k), (47k, 5k), and (118k, 5k). 
The object detection task requires predictions of bounding boxes and label predictions. 
It is evaluated by box average precision (AP). 
The keypoint detection task requires predictions of 17 human keypoints. 
It is evaluated by box AP and keypoint AP. 
The panoptic segmentation task combines instance segmentation and semantic labeling. 
It is evaluated by box AP, mask AP, and panoptic quality (PQ).

We compress the ``R50-FPN (3$\times$)'' models \cite{lin2017feature} from the Detectron2 Model Zoo\footnote{\url{https://github.com/facebookresearch/detectron2/blob/master/MODEL_ZOO.md}} for the three tasks.
The ResNet-50 backbone of the three models was initialized using the same ImageNet pretrained checkpoint and contains 23.4M parameters. The detection, keypoint detection, and panoptic segmentation models additionally involve task-specific architectures of 18.2M, 33.5M, and 22.5M parameters and are thus significantly larger than the ResNet-50 classification model.
We use a subset of the hyperparameters searched in ImageNet experiments for simplicity and nevertheless observe reasonably good results.


Like in CIFAR experiments, we implement unstructured magnitude pruning baselines using PyTorch's pruning API. 
As shown in \cref{fig:detectron}, our method significantly outperforms the baselines on all three models and all metrics, especially at high compression ratios. With the same performance loss, AQCompress is able to compress these models to up to 18$\times$ while iterative unstructured pruning achieves only 12$\times$ or less. These results confirm that the proposed AQCompress is applicable to off-the-shelf DNNs that are larger and trained for various tasks. 


\subsection{Pruning + AQCompress} \label{subsection:pr+aq}
\begin{figure}[t]
  \centering
  \includegraphics[width=0.38\textwidth]{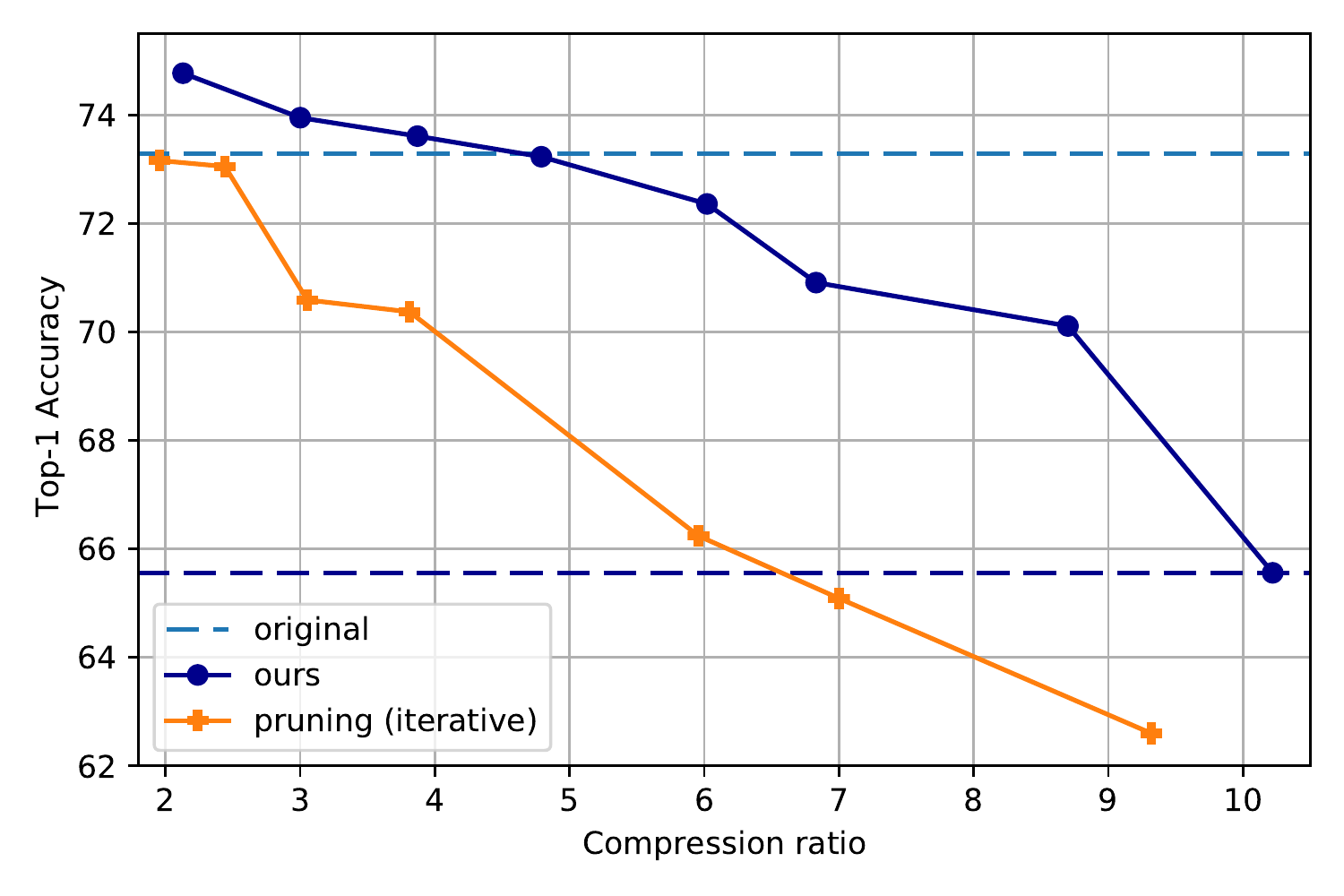}
  \caption{AQCompress for an iteratively pruned MobileNet-v2 with 20\% remaining parameters on CIFAR-100. The baseline continues to iteratively prune the pruned network. AQCompress remains effective on sparse networks and outperforms the baseline. }
  \label{p+aq}
\end{figure}

In this section, we further validate that AQCompress is architecture-agnostic by showing that AQCompress can compress not only parameters from dense architectures, but also those from sparse ones. We take an iteratively pruned MobileNet-v2 trained for CIFAR-100 with 20.97\% remaining parameters and form a parameter matrix for them. We continue the iterative pruning from this checkpoint to establish a baseline. We apply AQCompress to the parameter matrix and compare its size-accuracy trade-off curve against the one from continued iterative unstructured pruning in \cref{p+aq}. AQCompress clearly outperforms the baseline on parameters decoupled from the irregular network architecture. We thus suggest that AQCompress complements sparse network searching methods and can potentially be combined with other compression techniques to achieved high compression ratios.

\subsection{Ablation Study} \label{subsection:ablation}
\spara{Entire vs. Partial Network Additive Quantization}
Intuitively, allowing more parameters to share representations may lead to better size-accuracy trade-offs. In this section, we study whether learning shared representations for an entire DNN is more advantageous than compressing the model as multiple separate parts. 
We run a few experiments to compress MobileNet-v2 trained on CIFAR-100 to reach a $5.5\times$ compression ratio according to bit-wise code representations. 
We do not leverage Huffman coding in this experiment since the compression ratios resulting from it can not be predetermined. 
We pick the light-weight MobileNet-v2 on CIFAR-100 since it is a challenging pair for which we observe accuracy loss at high compression ratios for both pruning and AQCompress. 
We pick the $5.5\times$ compression ratio since the accuracy loss of AQCompress on compressing the entire model to this size is within 1\%. 
In each experiment, we evenly divide the model parameters into few groups based on their sequential order in the feedforward model architecture. 
We learn a set of code representations for each group. 
We set the parameter page size to 8 in all experiments. 
As shown in \cref{tab:part}, compressing the entire DNN results in the best task performance while learning 16 sets of code representations separately for 16 model parts incurs the most performance drop. 
The results support the assumption that compressing more layers together allows representation sharing across layers and thus lead to better size-accuracy trade-offs.

\begin{table}
\small
  \caption{Entire vs. partial network additive quantization. Compress MobileNet-v2 on CIFAR-100 to $5.5\times$ as an entire model achieves better top-1 accuracy than compressing it as multiple separate parts.}
  \label{tab:part}
  \centering
  \begin{tabular}{l|cccc}
    \toprule
    Number of Parts & 1 & 4 & 8 & 16  \\
    Top-1 Accuracy & 73.82 & 72.59 & 73.24 & 71.81 \\
    \bottomrule 
  \end{tabular}
\end{table}

\spara{Code Representations}
\cref{huffman} compares different possibilities for discrete code representations. Huffman coding yields more compact representations than bit-wise representations. Detailed descriptions of the bit-wise representation are in the Appendix. Sorting the basis vectors in codebooks further results in significantly more compression. We thus validate the advantage of our strategy of sorting basis vectors to achieve more skewed code distributions. 

\begin{table}
\footnotesize
  \caption{Compression ratios by bit-wise representations, Huffman coding alone, and Huffman coding following basis vector sorting. Huffman coding achieves better space-efficiency than bit-wise representations, while sorting basis vectors further improves it significantly. Example ResNet-50 checkpoints are from \cref{subsection:cls}.}
  \label{huffman}
  \centering
  \begin{tabular}{l|cccccc}
    \toprule
    Code representation & 
    \multicolumn{6}{c}{Compression ratios} \\
    \hline
    Bit-wise & 2.00 & 3.99 & 6.21 & 8.16 & 9.64 & 11.78 \\
    Huffman & 2.58 & 4.56 & 6.26 & 7.04 & 10.15 & 12.30 \\
    Sorting+Huffman & 4.79 & 7.19 & 9.97 & 11.49 & 12.97 & 15.27 \\
    \bottomrule
  \end{tabular}
\end{table}


\section{Analysis: Basis Vector Sharing and Usage Patterns}
In this section, we demonstrate the basis vector sharing patterns with the $8\times$ compression checkpoint of ResNet-50 trained on ImageNet classification. The hyperparameters are parameter page size $D=32$, $M=13$, $K=512$. Other checkpoints share the same patterns.

\spara{Sharing across parameter blocks} 
In this analysis, we aim to verify whether multiple network blocks naturally learn to share basis vectors. 
In all our experiments, the codebooks and discrete codes are randomly initialized and learned. 
We do not add any regularization about whether parameter pages from different network layers should share representations. 
We consider six parameter blocks in ResNet-50, including the input convolutional layer, the 4 residual blocks, and the output linear layer. 
For the $i^{th}$ parameter group, we collect the basis vectors (indexed $1$ to $MK$) needed to reconstruct its parameter pages
as a multiset $G_i$. 
We compute the sharing factor between two parameter groups as:
\begin{equation}
Sharing(G_i, G_j) = \frac{|Intersection (G_i, G_j)|}{ min(|G_i|, |G_j|)}.
\end{equation}

\begin{figure}[t]
  \centering
  \includegraphics[width=0.35\textwidth]{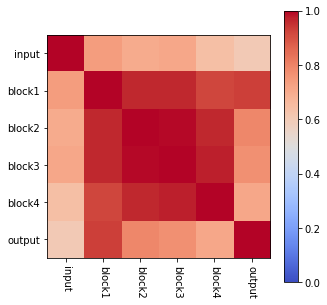}
  \caption{Basis vector sharing pattern analysis using the 8$\times$ compressed ResNet-50 by AQCompress. We observe larger than 0.6 pair-wise sharing factors among all network blocks.}
  \label{fig:analysis-block}
\end{figure}

Intuitively, the more overlaps between groups, the larger this sharing factor is. 
If a multiset is a subset of another,
the two groups have a sharing factor of 1, suggesting a strong sharing pattern. 
We plot the pair-wise sharing factors among the six parameter groups in Figure \ref{fig:analysis-block}. All sharing factors are above 0.6. 
The result is a strong evidence that parameter representation sharing frequently happens across network layers. 

\begin{figure}[t]
  \centering
  \includegraphics[width=0.47\textwidth]{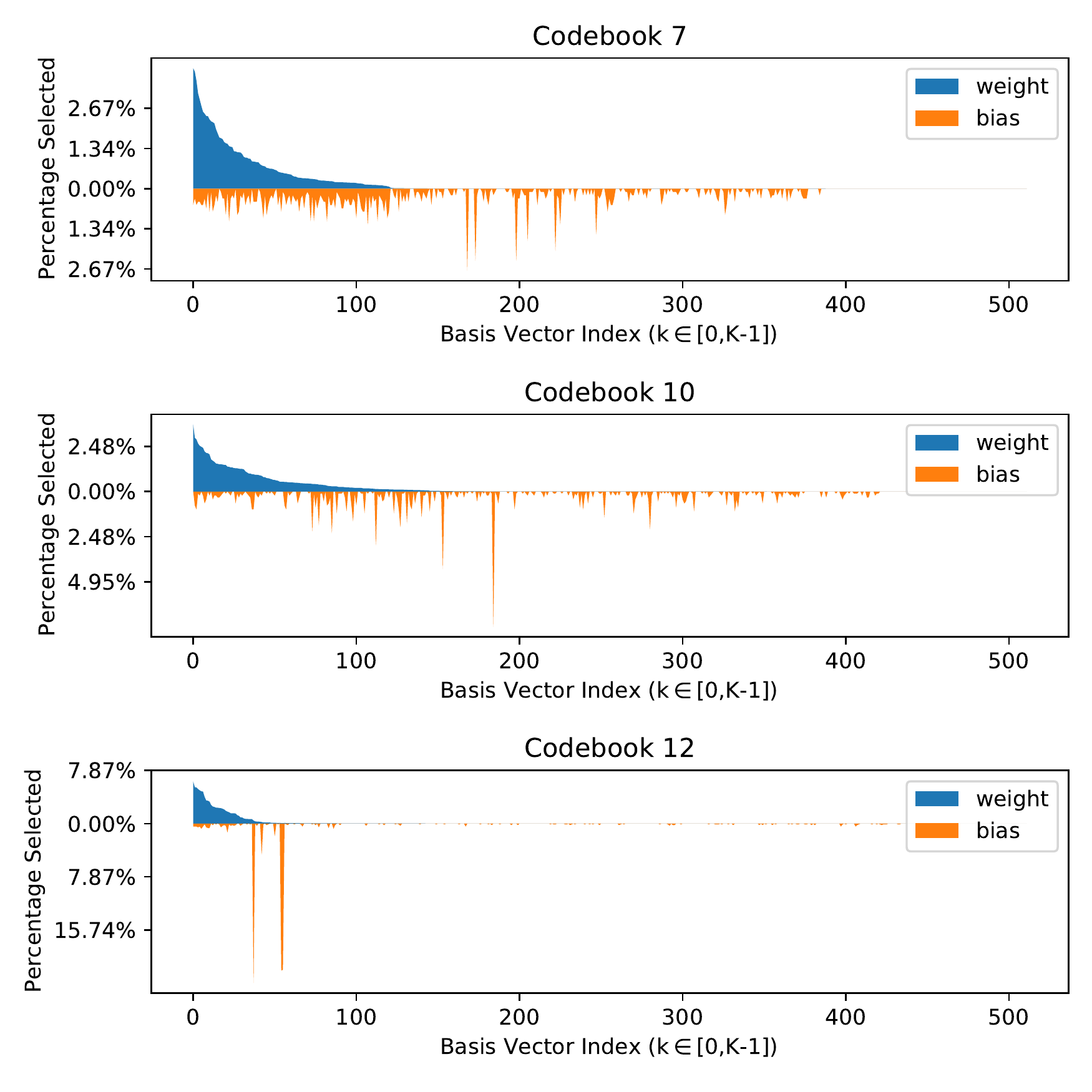}
  \vspace{-.2cm}
  \caption{Basis vector sharing within and between parameter types (weights vs. biases). For each basis vector, we plot in blue (yellow) the percentage of parameter pages involving weight (bias) parameters that use this basis vector. We observe that frequent basis vector sharing within and between the two types.}
  \label{fig:analysis-type}
  \vspace{-.2cm}
\end{figure}

\spara{Sharing within and between weight parameters and bias parameters}
We group the DNN parameters into weights and biases and examine degree to which parameter pages used by weights and biases share basis vectors within and between the two types. 
Within a codebook, for each basis vector, we plot the percentage of parameter pages involving weight (bias) parameters that use this basis vector in blue (orange) (\cref{fig:analysis-type}). The basis vectors are sorted by total usage frequencies. In all three codebooks, there are a number of basis vectors that make significant contributions to both weights and biases. In codebook 7, for example, about 30\% of the basis vectors contributes to 99.6\% of the parameters used for weights and 84\% of the parameters used for biases. This indicates that parameter pages are frequently shared within and between the two parameter types. 

\spara{Basis Vector Usage}
In some codebooks, Codebook 12 in \cref{fig:analysis-type} for example, the basis vector usage distribution is very skewed. Approximately 10\% of the basis vectors contribute to the additive composition of 94.9\% of parameter pages. We suggest that AQCompress can possibly be combined with mixed-precision representations to achieve higher compression ratios.

\section{Conclusion}

This paper investigates the possibility of compressing DNN parameters independently of their network architecture.
We propose an architecture-agnostic DNN compression scheme, AQCompress, that leverages additive quantization and Huffman coding to compactly represent and store trained DNN parameters.
The scheme can be widely applied to compress diverse models, including light-weight MobileNet-v2,  pruned networks, and so forth, trained for classification, detection, and segmentation tasks.
Extensive empirical results show that AQCompress is both competitive and complementary to competing compression methods such as iterative unstructured pruning. 
Further analyses show that AQCompress can learn compact neural representations shared across network layers in DNNs. 
We hope this paper inspires future research on DNN compression from the perspective of efficiently representing and storing DNN parameters, besides the efforts of finding and training light-weight DNN architectures.

Future work includes co-design of compression algorithms and hardware-specific system scheduling to support real-time inference. Parallelizing the decoding of codebook representations and the forward propagation of  reconstructed network layers may accelerate model inference. From the algorithm perspective, it remains an open question how to best learn the compact representations for DNN parameters. Promising techniques might combine lossy compression with DNN-specific compression techniques or explore more lossy compression methods other than additive quantization.

{\small
\bibliographystyle{ieee_fullname}
\bibliography{egbib}
}

\end{document}